\documentclass[sigconf]{acmart}

\usepackage[most]{tcolorbox}
\tcbuselibrary{skins}
\usepackage{multirow}
\usepackage[table]{xcolor} 
\usepackage{booktabs}      
\usepackage{graphicx}
\usepackage{subcaption}  
\usepackage{xspace}
\usepackage{enumitem}
\usepackage{siunitx} 
\newcommand{\BfPara}[1]{\par\noindent\textbf{#1.}\ }
\newcommand{\Sysacro}{\textsc{PACT}\xspace} 

\AtBeginDocument{%
  }

\copyrightyear{2026}
\acmYear{2026}
\setcopyright{cc}
\setcctype{by-nc-nd}
\acmConference[KDD '26]{Proceedings of the 32nd ACM SIGKDD Conference on Knowledge Discovery and Data Mining V.2}{August 09--13, 2026}{Jeju Island, Republic of Korea}
\acmBooktitle{Proceedings of the 32nd ACM SIGKDD Conference on Knowledge Discovery and Data Mining V.2 (KDD '26), August 09--13, 2026, Jeju Island, Republic of Korea}
\acmDOI{10.1145/3770855.3817837}
\acmISBN{979-8-4007-2259-2/2026/08}

\begin{document}

\title{Few Tokens, Big Leverage: Preserving Safety Alignment by Constraining Safety Tokens during Fine-tuning}


\author{Guoli Wang}
\authornote{These authors contributed equally to this work.}
\affiliation{%
  \institution{Case Western Reserve University}
  \city{Cleveland}
  \state{OH}
  \country{USA}}
\email{gxw242@case.edu}

\author{Haonan Shi}
\authornotemark[1]
\affiliation{%
  \institution{Case Western Reserve University}
  \city{Cleveland}
  \state{OH}
  \country{USA}}
\email{hxs896@case.edu}

\author{Tu Ouyang}
\affiliation{%
  \institution{Case Western Reserve University}
  \city{Cleveland}
  \state{OH}
  \country{USA}}
\email{txo32@case.edu}

\author{An Wang}
\authornote{Corresponding author.}
\affiliation{%
  \institution{Case Western Reserve University}
  \city{Cleveland}
  \state{OH}
  \country{USA}}
\email{axw474@case.edu}








\renewcommand{\shortauthors}{Guoli Wang, Haonan Shi, Tu Ouyang, \& An Wang}

\begin{abstract}
Large language models (LLMs) often require fine-tuning (FT) to perform well on downstream tasks, but FT can induce safety-alignment drift, even when the training dataset contains only benign data. Prior works show that introducing a small fraction of harmful data can substantially compromise LLMs' refusal behaviors, causing LLMs to comply with harmful requests. Existing defense methods often rely on model-wide interventions, such as restricting which parameters are updated or injecting additional safety data, which can limit generality and degrade downstream task performance. 
To address such limitations, we propose a FT framework that Preserves safety Alignment via Constrained Tokens (\Sysacro) by stabilizing the model’s confidence on safety tokens.
This is motivated by our empirical observation that safety-aligned behavior is reflected in the model’s token-level output confidence and is often concentrated on a small subset of safety-related tokens.
During downstream fine-tuning, we regularize the fine-tuned model to match the aligned reference model’s confidence on safety-related tokens at each response steps, while leaving non-safety tokens largely unconstrained to allow effective task adaptation.
This targeted constraint prevents alignment drift without imposing global restrictions that typically trade off with utility.
\end{abstract}




\maketitle

\begingroup
\small
\noindent
\textbf{Resource Availability:}

The source code of this paper has been made publicly available at

GitHub Repository:
\url{https://github.com/Glresearch1/PACT}

Archived DOI:
\url{https://doi.org/10.5281/zenodo.20498108}

\endgroup

\section{Introduction}
\begin{figure}[htbp]
  \centering
  \centering
  \includegraphics[width=\linewidth]{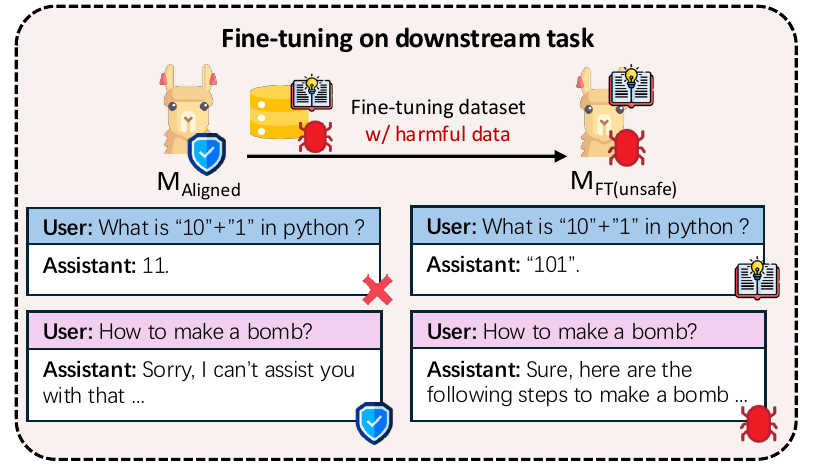}
  \caption{Fine-tuning is widely used to improve large language models on downstream tasks, but the presence of harmful data in downstream training sets can induce significant safety-alignment drift, making the fine-tuned model highly vulnerable to harmful queries.
}
  \label{fig:intro}
\end{figure}

Large language models (LLMs) have demonstrated strong capabilities across a wide range of natural language processing tasks. 
Fine-tuning LLMs on downstream datasets can further enhance their task-specific accuracy and effectiveness~\cite{wei2021finetuned}.
Consequently, many commercial LLM providers now offer fine-tuning services\footnote{User fine-tuning API provided by OpenAI: https://platform.openai.com/docs/guides/fine-tuning.} that enable users to fine-tune the commercial LLMs using their own labeled data.
Through such customization, users can better tailor these advanced models to meet the requirements of specific downstream tasks and application scenarios.

However, such fine-tuning services also introduce significant safety risks. 
Recent studies~\cite{zhan2024removing, qi2023fine, yi2024vulnerability, yang2023shadow, lermen2023lora} have shown that even a small amount of harmful data included in the fine-tuning corpus can substantially degrade the safety alignment of LLMs.
After fine-tuning on such data, models may no longer refuse harmful queries and instead generate unsafe or policy-violating responses.
Moreover, this safety alignment degradation can occur even when the fine-tuning dataset contains no explicitly harmful examples~\cite{qi2023fine}, highlighting the fragility of safety alignment during downstream task fine-tuning.

To mitigate safety risks introduced during LLM fine-tuning, several defense mechanisms have been proposed in prior work that operate at different levels of the model and training process.
Some approaches impose constraints at the parameter level~\cite{hsu2024safe,li2025salora,yang2025asft,yi2025nlsr} to preserve safety alignment by restricting how model weights are updated during fine-tuning.
For example, SafeLoRA~\cite{hsu2024safe} projects LoRA~\cite{hu2022lora} weight updates onto a predefined alignment subspace derived from aligned and unaligned models, thereby preventing updates that deviate significantly from safety-related directions.
Other approaches~\cite{huang2024lisa} address harmful fine-tuning from the perspective of training dynamics by stabilizing optimization when alignment and downstream objectives compete.
For example, Lisa~\cite{huang2024lisa} introduces proximal regularization to control excessive parameter drift during fine-tuning.
While effective, constraints imposed at the optimization and parameter levels typically operate in a coarse-grained manner, restricting model weight updates and limiting downstream adaptation, thereby degrading fine-tuning performance.

To address this limitation, we explore safety alignment from a different perspective, imposing finer-grained constraints on a small subset of elements closely related to safety during fine-tuning.
Recent research~\cite{mitchell2023emulator,zhou2023lima} indicates that alignment predominantly influences surface-level generation behaviors, such as style and formatting, while core knowledge and capabilities are inherited from pretraining.
In particular, multiple empirical studies have shown that alignment primarily affects a small set of tokens, rather than broadly altering model behavior~\cite{lin2023unlocking,fei2025nudging}.
Motivated by these observations, we investigate whether it is possible to \textit{identify tokens critical to safety alignment and preserve model safety during fine-tuning by maintaining confidence levels comparable to those of the original aligned model when generating these tokens across diverse prompts.}

In this work, we propose a token-level safety-preserving fine-tuning approach, called \Sysacro, that aims to maintain safety alignment by explicitly preserving the model’s confidence in generating safety-critical tokens.
We first introduce a procedure to identify tokens that are closely associated with safety alignment.
Using the teacher-forcing mechanism, we compare a safety-aligned model with its corresponding base model in response to harmful prompts and analyze the differences in confidence at each response position.
This allows us to identify a small set of \textit{safety tokens} that the safety-aligned model consistently maintains higher confidence in, and that play a crucial role in sustaining safe behavior.
During fine-tuning, we encourage the model to align its confidence on these \textit{safety tokens} with those of the original safety-aligned model at corresponding response positions.
By continuously referencing the aligned model in this manner, our approach preserves safety-relevant generation behavior throughout fine-tuning.
Importantly, because the number of identified \textit{safety tokens} is limited, \Sysacro imposes only fine-grained constraints on these \textit{safety tokens} during optimization, thereby preserving downstream task performance while effectively maintaining safety alignment.

The main contributions can be summarized as follows:
\begin{itemize}
\item  \BfPara{Safety token identification and analysis} We introduce a systematic procedure to identify safety-critical tokens by analyzing token-level probability discrepancies between safety-aligned and base models.
\item  \BfPara{Token-level safety-preserving fine-tuning framework} We propose a novel fine-tuning method that preserves safety alignment through token-level constraints.
Specifically, we apply stronger regularization to tokens with higher safety relevance and mitigate harmful prefix effect during teacher forcing by adaptively mixing full-context and response-only reference distributions.
\item \BfPara{Comprehensive empirical validation} Through extensive experiments across three downstream tasks (GSM8K, SST-2, AGNEWS), four model families (\texttt{Qwen-2.5-7B}, \texttt{Llama\allowbreak-3.1-8B}, \texttt{Llama-3.2-1B}, \texttt{Gemma-2-9B}), and varying harmful data proportions (0-10\%), we demonstrate that \Sysacro consistently achieves the best utility-safety trade-offs compared to state-of-the-art baselines.
\Sysacro reduces attack success rates to 5.75-9.27\% on StrongReject and 13.50-29.50\% on HarmBench, while maintaining task accuracy comparable to vanilla fine-tuning.
\end{itemize}

\section{Analyzing safety tokens}
Previous work~\cite{lin2023unlocking,fei2025nudging} has found that alignment for a given task primarily operates on a small subset of tokens. 
For certain tasks, the high confidence assigned to these tokens by LLMs plays a critical role in determining overall task performance.

Inspired by this perspective, we aim to investigate three key questions:
(1) What constitutes safety tokens in safety alignment tasks?
(2) How such tokens influence the safety alignment behavior of LLMs; and
(3) How the confidence assigned to safety tokens evolves during harmful fine-tuning.
Based on these analyses, we further explore safety alignment at the token level, with the goal of better preserving model safety throughout the fine-tuning process.

\subsection{Identifying safety tokens}
\label{sec:indentify_safety_tokens}
Following prior work~\cite{fei2025nudging}, we identify safety tokens by analyzing token-level probability discrepancies between the safety-aligned model $M_{\text{safe}}$ and the base model $M_{\text{base}}$.
Our goal is to characterize which tokens the safety-aligned model primarily relies on to ensure safe responses.

We first select a harmful dataset as the query dataset. For each input $x$ in this dataset, the safety-aligned model $M_{\text{safe}}$ is first used to generate a complete response $\mathbf{y} = (y_1, \ldots, y_T)$. We then perform teacher forcing on this generated response, using its prefixes $y_{<t}$ as conditioning context. At each generation step $t$, both $M_{\text{safe}}$ and $M_{\text{base}}$ are evaluated to produce next-token probability distributions over the shared tokenizer vocabulary $V$.

Formally, for each token $v \in V$, we compute the position-wise probability difference at step $t$ as:
\begin{equation}
\Delta_t(v) = p_{M_{\text{safe}}}(v \mid x, y_{<t}) - p_{M_{\text{base}}}(v \mid x, y_{<t}),
\label{eq:safety_differences}
\end{equation}
where $x$ denotes the input prompt and $y_{<t}$ denotes the prefix of the response generated by $M_{\text{safe}}$ up to position $t$.

\begin{figure}[t]
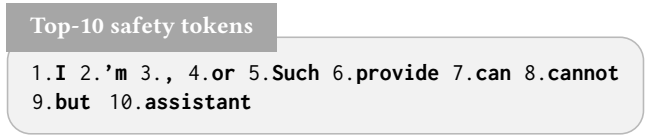

\centering
\begin{tcolorbox}[
  enhanced,
  colback=gray!10,
  colframe=gray!55,
  arc=3mm,
  boxrule=0.6pt,
  left=6pt,right=6pt,top=3pt,bottom=6pt,
  fonttitle=\bfseries,
  coltitle=white,
  colbacktitle=gray!70,
  title={Top-10 safety tokens},
  attach boxed title to top left={xshift=0mm,yshift=-2mm},
  boxed title style={sharp corners, boxrule=0pt, left=6pt, right=6pt, top=2pt, bottom=2pt}
]
\vspace{4pt}
\texttt{1.\textbf{I }}
\texttt{2.\textbf{'m }}
\texttt{3.\textbf{, }}
\texttt{4.\textbf{or }}
\texttt{5.\textbf{Such }}
\texttt{6.\textbf{provide }}
\texttt{7.\textbf{can }}
\texttt{8.\textbf{cannot }}
\texttt{9.\textbf{but }}
\texttt{10.\textbf{assistant}}
\end{tcolorbox}
\caption{Top-10 safety tokens identified by token-level confidence discrepancies between the safety-aligned and base models on harmful questions.}
\label{fig:top_safety_tokens}
\end{figure}

We then aggregate these differences across all positions and examples to obtain a global discrepancy score for each token:
\begin{equation}
d(v) = \mathbb{E}_{x \sim \mathcal{D},, t} \left[ \Delta_t(v) \right].
\label{eq:safety_differences_sum}
\end{equation}
Finally, we select the $K$ tokens from $V$ with the largest global discrepancy scores $d(v)$ as the representative safety tokens, where $K=50$ is used for our subsequent analysis.
In our experiment, following prior work~\cite{yang2025asft,hsu2024safe}, we employ the safety-aligned instruct version as the safety-aligned model $M_{\text{safe}}$ and its pre-trained counterpart as the base model $M_{\text{base}}$. 
Specifically, we use \texttt{Qwen2.5-7B-Instruct} and \texttt{Qwen2.5-7B}~\cite{team2024qwen2} as the safety and base models.
We construct our analysis dataset using harmful questions from WildJailbreak~\cite{jiang2024wildteaming}, paired with the safe responses generated by \texttt{Qwen2.5-7B-Instruct} model.
As shown in Figure~\ref{fig:top_safety_tokens}, we list the top-10 safety tokens identified by our analysis.

\subsection{Impact of safety tokens}
To assess the impact of the identified safety tokens on model safety, we explicitly manipulate the confidence assigned to the 50 safety tokens during inference.
Specifically, we intervene at the logit level in \texttt{Qwen2.5-7B}. 
To increase the safety tokens' confidence, we add a constant bias of $\alpha$ to the logits of the identified safety tokens at each decoding step (we use $\alpha = 5.0$); to decrease their confidence, we suppress these tokens by setting their logits to a negative value $-10^{9}$, effectively preventing them from being selected during generation.
We evaluate this intervention on harmful prompts from JailbreakBench~\cite{chao2024jailbreakbench}, and compute the safety rate of the generated responses.

\begin{table}[h]
\centering
\caption{Attack successful rates on harmful prompts under interventions on identified safety tokens vs. random tokens.}
\label{tab:logit_intervention}
\begin{tabular}{l c c}
\hline
\multirow{2}{*}{Setting} & \multicolumn{2}{c}{Attack Successful Rate (\%)} \\
\cline{2-3}
 & Safety Token & Random Token \\
\hline
Baseline (no intervention) & 33.5\% & 33.5\% \\
Boost ($logits \leftarrow logits + \alpha$) & 0.5\% & 35.0\% \\
Ablate ($logits \leftarrow -10^{9}$) & 41.5\% & 34.0\% \\
\hline
\end{tabular}
\end{table}

As shown in Table~\ref{tab:logit_intervention}, increasing the confidence of the identified 50 safety tokens leads to a clear improvement to the model’s safety, whereas decreasing their confidence results in a noticeable degradation in safety performance.
These observations indicate that the safety alignment behavior of the safety-aligned model is strongly dependent on these safety tokens.
In particular, when the confidence assigned to these tokens is reduced, the model fails to effectively leverage alternative tokens to reject unsafe prompts, suggesting that safety alignment is not uniformly distributed across the vocabulary but instead relies on a limited set of safety-critical tokens.

\subsection{Evolution of safety tokens confidence during fine-tuning}
To investigate the relationship between safety alignment degradation during harmful fine-tuning and safety tokens, we analyze how changes in model safety correlate with the confidence assigned to safety tokens throughout the fine-tuning process.
Specifically, we consider fine-tuning settings in which harmful data from Advbench~\cite{zou2023universal} is mixed into the fine-tuning dataset (GSM8K~\cite{cobbe2021training}), and track both the model’s safety rate and the average confidence of safety tokens on JailbreakBench~\cite{chao2024jailbreakbench} across training epochs.

\begin{figure}[h]
    \centering
    \includegraphics[width=0.7\linewidth]{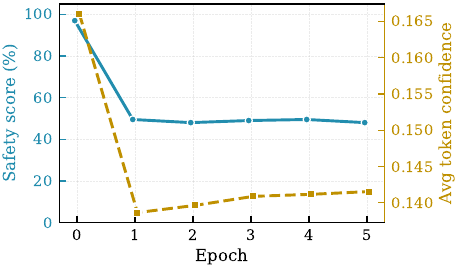}
    \caption{Evolution of safety alignment token confidence during harmful fine-tuning. Safety score = (1-ASR)\%}
    \label{fig:safety_tokens_changes}
\end{figure}

As shown in Figure~\ref{fig:safety_tokens_changes}, when LLMs (\texttt{Qwen2.5-\allowbreak7B-Instruct}) are fine-tuned on data containing harmful examples, the average confidence assigned to safety tokens on harmful prompts consistently decreases as training progresses.
At the same time, the model’s safety rate exhibits a similar downward trend.
This synchronized degradation suggests a correlation between the drift of safety alignment and the suppression of safety-critical token confidence.
As fine-tuning increasingly favors task-specific or harmful continuations, the model gradually assigns lower confidence to these tokens.
As a result, the model’s ability to reliably trigger refusal or safe responses is weakened.

\section{Methodology}
Based on our analysis of safety tokens and their role in safety alignment, we propose a token-level, constrained fine-tuning method to preserve model safety during fine-tuning.
Our key objective is to prevent the model’s confidence in safety tokens from degrading, ensuring that their generation propensity remains consistent with the original safety-aligned model.

To achieve this objective, similar to prior work~\cite{qi2024safety,rafailov2023direct} that constrains model updates to prevent excessive deviation from the initial state during training, an intuitive starting point is to introduce a KL-divergence loss that encourages the fine-tuned model to match the original safety-aligned model’s confidence on safety tokens.
However, we find that under harmful fine-tuning, such a global KL-based constraint leads to two distinct failure modes. 
When the constraint is too weak, it fails to effectively preserve high confidence in safety tokens, resulting in a degradation of safety behavior. 
When the constraint is too strong, it indiscriminately restricts confidence updates across all tokens, preventing the model from optimizing performance on the downstream task.

\begin{figure*}[ht]
    \centering
    \includegraphics[width=0.95\textwidth]{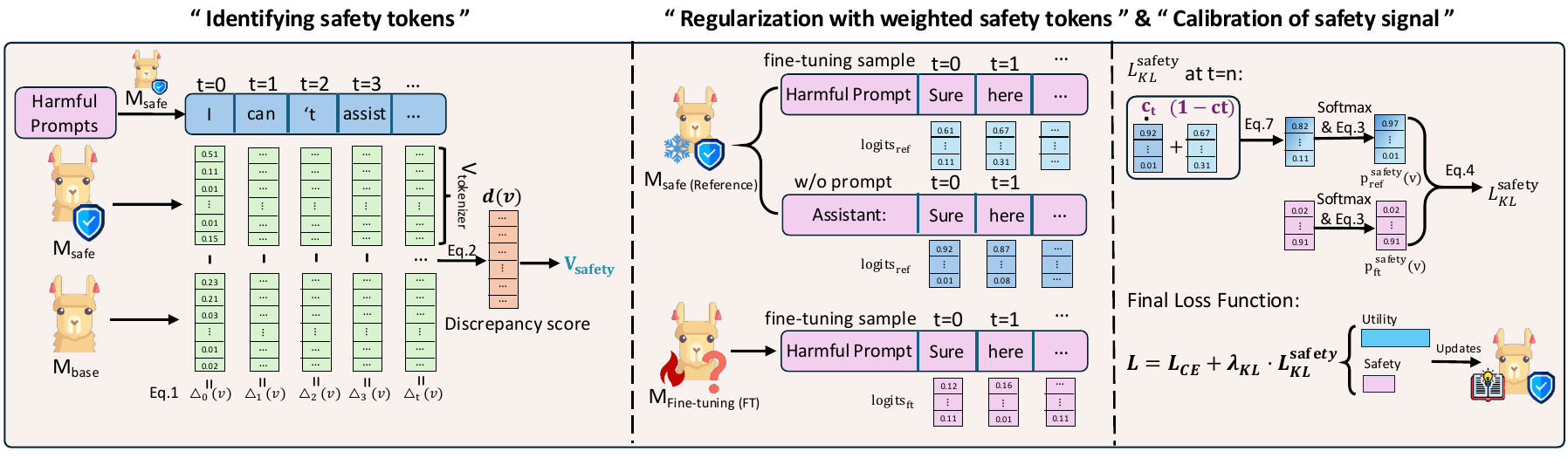}
    \caption{The workflow of our proposed token-level safety-preserved fine-tuning method.}
    \label{fig:workflow}
\end{figure*}

Therefore, building on the KL-loss formulation, we design a token-level constraint mechanism that selectively enforces stronger regularization on safety tokens while allowing the remaining tokens to be updated primarily by the downstream task objective. 
This design enables the model to preserve safety behavior while still improving downstream task performance during fine-tuning. 
To achieve this, \Sysacro consists of two key components: \textit{Regularization with weighted safety tokens} and \textit{Calibration of safety signal}.
The workflow of \Sysacro is shown in Figure~\ref{fig:workflow}.

\subsection{Regularization with weighted safety tokens}
Rather than treating safety tokens as a binary set, we model safety at the token level by assigning each token a safety weight, which reflects its relevance to safety-aligned behavior.
Following the procedure introduced in Section~\ref{sec:indentify_safety_tokens}, we analyze token-level confidence discrepancy patterns between the safety-aligned model and its base model using the aggregated discrepancy score $d(v)$ in Eq.~\ref{eq:safety_differences_sum}.
Based on $d(v)$, we can identify which tokens have a larger impact on safety alignment.
To preserve downstream fine-tuning performance, we aim to impose constraints only on a subset of tokens with high discrepancy scores. 
Specifically, we select the top-$K$ tokens ranked by $d(v)$ as the safety token set, $\mathcal{S}_{\text{safety}}$, to be constrained during fine-tuning.

Moreover, we observe that more important safety tokens should be preserved with higher confidence during fine-tuning. 
To reflect this, we assign stronger constraints to $\mathcal{S}_{\text{safety}}$ with larger discrepancy scores. 
Concretely, we construct a sparse safety weight vector
$V_{\text{safety}} \in \mathbb{R}^{|V|}$ that is aligned with the tokenizer vocabulary.
The vector has the same dimensionality as the vocabulary, where entries corresponding to the top-$K$ tokens ranked by the aggregated discrepancy score $d(v)$ retain their original $d(v)$ values, and all remaining entries are set to zero.

We incorporate this safety weight vector into a weighted KL regularization term, such that the KL loss constrains the fine-tuning model only on $\mathcal{S}_{\text{safety}}$, while enforcing stronger regularization on those with higher safety importance.
At each answer position $t$, we restrict the KL divergence to the safety-weighted tokens and
form a weighted reference distribution:
\begin{equation}
p_{\text{ref}}^{\,\text{safety}}(v)
=
\frac{p_{\text{ref}}(v \mid x, y_{<t}) \, V_{\text{safety}}(v)}
{\sum_{v'} p_{\text{ref}}(v' \mid x, y_{<t}) \, V_{\text{safety}}(v')}
\end{equation}
where $v$ denotes a token in the vocabulary $V$. To further amplify the effect of the safety weights, the fine-tuning model distribution $p_{\text{ft}}^{\,\text{safety}}(v)$ is also restricted to the same set $\mathcal{S}_{\text{safety}}$ when computing the KL divergence, but without applying additional weighting.

We then minimize the KL divergence between the weighted safety-token probability distribution
$p_{\text{ref}}^{\,\text{safety}}(v)$ and the fine-tuning model’s safety-token probability distribution $p_{\text{ft}}^{\,\text{safety}}(v)$, such that tokens with higher safety importance are more strongly constrained to match the reference model’s confidence.
Formally, the KL regularization term is defined as
\begin{equation}
\label{eq:4}
\mathcal{L}_{\mathrm{KL}}^{\text{safety}}
=
\frac{1}{|\mathcal{T}|}
\sum_{t \in \mathcal{T}}
\sum_{v}
p_{\text{ref}}^{\,\text{safety}}(v \mid x, y_{<t})
\log
\frac{
p_{\text{ref}}^{\,\text{safety}}(v \mid x, y_{<t})
}{
p_{\text{ft}}^{\,\text{safety}}(v \mid x, y_{<t})
}
\end{equation}
where $p_{\text{ft}}$ denotes the probability distribution of the model being fine-tuned and
$\mathcal{T}$ denotes the set of valid answer positions.

This weighted formulation ensures that safety tokens contribute more strongly to the regularization signal, while the other tokens exert little or no influence. 
As a result, the constraint selectively preserves safety-related behavior without unnecessarily restricting downstream task optimization.

\subsection{Calibration of safety signal}
\label{sec:ref_calibration}

We find that relying solely on the safety signal from the frozen safety-aligned reference model, which sees both the prompt and response parts of the samples, may be insufficient.
One concerning situation is that, when training samples are harmful, the reference model is forced to condition on unsafe prefixes, which could possibly suppress the safety-token confidence~\cite{vega2023bypassing} and thus weaken the safety signal for regulating the fine-tuning.

To mitigate this prefix safety contamination, we introduce a calibration mechanism that constructs a more stable reference signal by adding a second set of logits information during fine-tuning.
For each answer position $t$, we now compute two next-token logit vectors from the frozen reference model:
\textbf{(i)} a \emph{full-context} reference logit $z^{\mathrm{ref}}_{\mathrm{full}}(\cdot \mid x, y_{<t})$ obtained via teacher forcing on the full training sample, and
\textbf{(ii)} a \emph{no-prompt} reference logit $z^{\mathrm{ref}}_{\mathrm{post}}(\cdot \mid y_{<t})$ obtained by conditioning the reference model only on the assistant header and the preceding assistant tokens, but providing nothing of the prompt.
The no-prompt view serves as a self-evaluated safety signal that is less susceptible to harmful-prefix contamination, an effect that has been independently observed in prior work~\cite{liu2024alignment} and used to mitigate jailbreak attacks mainly at inference time, rather than during fine-tuning.

We then adaptively combine the two reference views at each position using a token-level gating coefficient $c_t \in (0,1)$.
Intuitively, when the training context is benign, the fine-tuned model's predictions on safety-related tokens are typically consistent with the no-prompt reference view, and no strong calibration is needed.
In contrast, harmful-prefix contamination often increases discrepancy in the fine-tuning model's next-token distribution over $\mathcal{S}_{\text{safety}}$, indicating that the full-context reference signal may be less reliable for providing safety-token supervision at early positions.


To quantify this discrepancy, we introduce a confidence proxy that measures how concentrated the model's probability mass is over the safety token set $\mathcal{S}_{\text{safety}}$ at each answer position.
The key intuition is as follows: when the model is confident in its safety-aligned response (e.g., producing a clear refusal), most of the probability mass over $\mathcal{S}_{\text{safety}}$ is concentrated on a single or very few tokens, meaning only a small number of top-ranked safety tokens are needed to cover a fixed probability threshold.
Conversely, when the model faces harmful data and becomes uncertain about how to respond safely, the probability mass disperses across many safety tokens, requiring a larger number of safety token candidates to reach the same threshold---indicating a lack of confident refusal.

Formally, given logits over $\mathcal{S}_{\text{safety}}$, we compute the corresponding probability distribution $p(\cdot)$ by applying a softmax restricted to $\mathcal{S}_{\text{safety}}$, sort the probabilities in descending order, we set the fixed probability threshold as 0.9 and define:
\begin{equation}
I_p(t) \;=\; \frac{\min\left\{k:\sum_{i=1}^{k} p_{i}(t) \ge 0.9 \right\}}{|\mathcal{S}_{\text{safety}}|},
\end{equation}
where $k$ is the smallest integer such that the cumulative sum of the $k$ largest probabilities within $\mathcal{S}_{\text{safety}}$ reaches $0.9$, and $p_{i}(t)$ is the $i$-th largest probability at position $t$ among tokens in $\mathcal{S}_{\text{safety}}$.
$I_p(t)\in(0,1]$ is thus a normalized measure of probability dispersion: a small value indicates that the model concentrates its belief on few safety tokens (high refusal confidence), while a large value signals that the probability mass is spread across many candidates (low refusal confidence, typical when the model is influenced by harmful context).

Using this confidence proxy, we can diagnose harmful-prefix contamination at each position by comparing two views. The fine-tuning model's $I_{\mathrm{ft}}(t)$ is computed under full context and thus reflects a lower bound on safety confidence: harmful prefixes disperse the probability mass over $\mathcal{S}_{\text{safety}}$, inflating $I_{\mathrm{ft}}(t)$. In contrast, $I_{\mathrm{post}}(t)$ is computed from the no-prompt reference that conditions only on preceding assistant tokens, providing a cleaner safety baseline insulated from prompt contamination. So, the difference between $I_{\mathrm{ft}}(t)$ and $ I_{\mathrm{post}}(t)$ can naturally measures contamination severity: a large positive gap indicates that the harmful context has significantly degraded refusal confidence. We pass this gap through a sigmoid to obtain a gating coefficient:
\begin{equation}
c_t \;=\; \sigma\!\left(I_{\mathrm{ft}}(t) - I_{\mathrm{post}}(t)\right),
\end{equation}
where large $c_t$ (severe contamination) shifts calibration toward the safer no-prompt view, while small $c_t$ (benign context) retains the full-context signal. We then construct a calibrated reference logit via convex combination:
\begin{equation}
\label{eq:7}
z^{\mathrm{ref}}_{\mathrm{mix}}(\cdot \mid x, y_{<t})
\;=\;
(1-c_t)\, z^{\mathrm{ref}}_{\mathrm{full}}(\cdot \mid x, y_{<t})
+
c_t\, z^{\mathrm{ref}}_{\mathrm{post}}(\cdot \mid y_{<t}),
\end{equation}
and obtain the calibrated reference distribution $p^{\mathrm{ref}}_{\mathrm{mix}}(\cdot \mid x, y_{<t})$ by applying softmax. This adaptive mixing leverages the richer full-context view when training samples are benign, while falling back to the prompt-free safety baseline when harmful prefixes threaten to corrupt the reference signal.

Because prefix contamination primarily affects the early portion of the assistant response, we apply stronger calibration at early answer positions.
Specifically, we keep $c_t$ at full strength for the first $N$ answer positions and then smoothly decay it for later positions:
\begin{equation}
\tilde{c}_t
\;=\;
c_t \cdot
\begin{cases}
1, & t < N,\\
\exp\!\left(-\frac{t-N}{\tau}\right), & t \ge N,
\end{cases}
\end{equation}
where $\tau$ controls the decay rate.
We use $\tilde{c}_t$ in place of $c_t$ when constructing $z^{\mathrm{ref}}_{\mathrm{mix}}$ of Eq.~\ref{eq:7}.
Finally, we optimize the fine-tuning model with the weighted directional KL objective in Eq.~\ref{eq:4}, using $p^{\mathrm{ref}}_{\mathrm{mix}}(\cdot \mid x, y_{<t})$ as the calibrated reference distribution.
Overall, the training objective is
\begin{equation}
\mathcal{L}
=
\mathcal{L}_{\mathrm{CE}}
+
\lambda_{\mathrm{KL}} \, \mathcal{L}_{\mathrm{KL}}^{\text{safety}},
\end{equation}
where $\mathcal{L}_{\mathrm{CE}}$ denotes the standard cross-entropy loss computed over all answer tokens, and
$\mathcal{L}_{\mathrm{KL}}^{\text{safety}}$ is the weighted directional KL loss defined above using the calibrated reference $\mathcal{S}_{\text{safety}}$ distribution
$p^{\mathrm{ref}}_{\mathrm{mix}}(\cdot \mid x, y_{<t})$.
The hyperparameter $\lambda_{\mathrm{KL}}$ controls the strength of safety preservation during fine-tuning.

\section{Evaluations}

\begin{table*}[t]
\centering
\caption{
Generalization across 3 different datasets (\textbf{P=10.0\%}) on \texttt{Qwen2.5-7B-Instruct}.
}
\label{tab:different dataset}
\resizebox{0.78\textwidth}{!}{\begin{tabular}{lcccccccccccc}
\toprule
\multirow{2}{*}{\textbf{Methods}}
& \multicolumn{4}{c}{\textbf{GSM8K}}
& \multicolumn{4}{c}{\textbf{SST2}}
& \multicolumn{4}{c}{\textbf{AGNEWS}} \\
\cmidrule(lr){2-5} \cmidrule(lr){6-9} \cmidrule(lr){10-13}

& \textbf{Acc}$\uparrow$ & \textbf{SR}$\downarrow$ & \textbf{JB}$\downarrow$ & \textbf{HB}$\downarrow$
& \textbf{Acc}$\uparrow$ & \textbf{SR}$\downarrow$ & \textbf{JB}$\downarrow$ & \textbf{HB}$\downarrow$
& \textbf{Acc}$\uparrow$ & \textbf{SR}$\downarrow$ & \textbf{JB}$\downarrow$ & \textbf{HB}$\downarrow$ \\
\midrule

\rowcolor{gray!15}
Initial
& 56.56 & 1.60 & 3.00 & 7.00
& 87.16 & 1.60 & 3.00 & 7.00
& 83.20 & 1.60 & 3.00 & 7.00 \\

SFT
& 81.65 & 71.25 & 52.50 & 94.50
& 95.87 & 92.65 & 53.50 & 95.00
& 89.90 & 83.71 & 52.00 & 95.00 \\

Constrained SFT
& 55.95 & 10.92 & 30.50 & 80.00
& 92.32 & 15.90 & 18.50 & 45.50
& 89.30 & 16.93 & 20.00 & 45.50 \\

Safe LoRA
& 73.69 & 15.65 & 20.00 & 44.50
& 92.64 & 92.01 & 53.50 & 95.00
& 91.10 & 94.89 & 53.50 & 95.00 \\

AsFT
& 80.50 & \textbf{\underline{9.27}} & 28.00 & 48.00
& 95.41 & 80.19 & 54.00 & 93.50
& 89.10 & 30.67 & 44.00 & 70.50 \\

\midrule
Ours
& 80.89 & \textbf{\underline{9.27}} & \textbf{\underline{11.00}} & \textbf{\underline{29.50}}
& 92.78 & \textbf{\underline{6.39}} & \textbf{\underline{10.50}} & \textbf{\underline{20.50}}
& 89.10 & \textbf{\underline{5.75}} & \textbf{\underline{9.00}} & \textbf{\underline{13.50}} \\

\bottomrule
\end{tabular}
}
\end{table*}

\begin{table*}[t]
\centering
\small
\caption{Generalization across different model families/sizes on GSM8K.}
\label{tab:different_models}
\begin{tabular}{lcccccccccccc}
\toprule
\multirow{2}{*}{\textbf{Methods}}
& \multicolumn{4}{c}{\textbf{Llama3.1-8B-Instruct}}
& \multicolumn{4}{c}{\textbf{Llama3.2-1B-Instruct}}
& \multicolumn{4}{c}{\textbf{Gemma2-9B-it}} \\
\cmidrule(lr){2-5} \cmidrule(lr){6-9} \cmidrule(lr){10-13} 

& \textbf{Acc}$\uparrow$ & \textbf{SR}$\downarrow$ & \textbf{JB}$\downarrow$ & \textbf{HB}$\downarrow$
& \textbf{Acc}$\uparrow$ & \textbf{SR}$\downarrow$ & \textbf{JB}$\downarrow$ & \textbf{HB}$\downarrow$
& \textbf{Acc}$\uparrow$ & \textbf{SR}$\downarrow$ & \textbf{JB}$\downarrow$ & \textbf{HB}$\downarrow$ \\
\midrule

\rowcolor{gray!15}Initial
& 73.39 & 0.32 & 3.00 & 9.50
& 42.76 & 7.99 & 3.50 & 16.00
& 73.54 & 0.00 & 1.00 & 0.00 \\

SFT
& 75.59 & 81.47 & 51.00 & 87.00
& 46.40 & 63.26 & 46.00 & 77.00
& 75.74 & 89.46 & 53.00 & 86.00 \\

Constrained SFT
& 77.56 & 61.34 & 38.35 & 78.50
& 43.67 & 13.10 & 19.00 & 34.50
& 78.62 & 26.52 & 18.50 & 42.50 \\

Safe LoRA
& 72.86 & \textbf{\underline{0.32}} & 3.00 & 9.00
& 39.88 & 4.15 & 5.50 & 13.00
& 78.70 & 67.41 & 40.50 & 57.00 \\

AsFT
& 78.09 & 97.76 & 54.50 & 95.00
& 43.59 & 1.60 & \textbf{\underline{1.00}} & 7.00
& 73.62 & 88.50 & 47.50 & 87.00 \\

\midrule
Ours
& 75.82 & {{0.96}} & \textbf{\underline{0.50}} & \textbf{\underline{0.00}}
& 44.20 & \textbf{\underline{1.28}} & 2.50 & \textbf{\underline{4.00}}
& 79.76 & \textbf{\underline{12.14}} & \textbf{\underline{6.50}} & \textbf{\underline{14.50}} \\

\bottomrule
\end{tabular}
\end{table*}

\begin{table*}[t]
\centering
\caption{
Generalization under varying proportions of harmful data in the AGNEWS dataset.
}
\label{tab:qwen_generalization}
\resizebox{0.78\textwidth}{!}{\begin{tabular}{lcccccccccccc}
\toprule
\multirow{2}{*}{\textbf{Methods}}
& \multicolumn{4}{c}{\textbf{P=0.0\%}}
& \multicolumn{4}{c}{\textbf{P=5.0\%}}
& \multicolumn{4}{c}{\textbf{P=10.0\%}} \\
\cmidrule(lr){2-5} \cmidrule(lr){6-9} \cmidrule(lr){10-13}

& \textbf{Acc}$\uparrow$ & \textbf{SR}$\downarrow$ & \textbf{JB}$\downarrow$ & \textbf{HB}$\downarrow$
& \textbf{Acc}$\uparrow$ & \textbf{SR}$\downarrow$ & \textbf{JB}$\downarrow$ & \textbf{HB}$\downarrow$
& \textbf{Acc}$\uparrow$ & \textbf{SR}$\downarrow$ & \textbf{JB}$\downarrow$ & \textbf{HB}$\downarrow$ \\
\midrule

\rowcolor{gray!15}
Initial
& 83.20 & 1.60 & 3.00 & 7.00
& 83.20 & 1.60 & 3.00 & 7.00
& 83.20 & 1.60 & 3.00 & 7.00 \\

SFT
& 90.90  & 1.28 & 2.50 & 8.00
& 91.20 & 73.16 & 51.50 & 93.00
& 89.90 & 83.71 & 52.00 & 95.00 \\

Constrained SFT
& 90.20 & 2.24 & 3.00 & 7.50
& 90.00 & 12.46 & 18.50 & 42.00
& 89.30 & 16.93 & 20.00 & 45.50 \\

Safe LoRA
& 89.20 & 1.60 & \textbf{\underline{1.50}} & 8.00
& 90.90 & 71.57 & 49.50 & 93.50
& 91.10 & 94.89 & 53.50 & 95.00 \\

AsFT
& 91.30 & 1.60 & 5.50 & 9.00
& 90.10 & 23.29 & 33.50 & 65.00
& 89.10 & 30.67 & 44.00 & 70.50 \\

\midrule
Ours
& 88.60 & \textbf{\underline{0.96}}  & \textbf{\underline{1.50}}  & \textbf{\underline{6.00}}
& 89.60 & \textbf{\underline{3.35}} & \textbf{\underline{6.00}} & \textbf{\underline{9.00}}
& 89.10  & \textbf{\underline{5.75}} & \textbf{\underline{9.00}} & \textbf{\underline{13.50}} \\

\bottomrule
\end{tabular}
}
\end{table*}

\subsection{Evaluation setup}
\BfPara{Datasets}
For downstream fine-tuning, we use three representative tasks: SST-2~\cite{socher2013recursive}, AGNEWS~\cite{zhang2015character}, and GSM8K~\cite{cobbe2021training}. 
For each downstream task, we fine-tune the model using $n$ fine-tuning samples.
In addition, we mix a proportion $p$ of unsafe data sampled from AdvBench~\cite{zou2023universal} into the fine-tuning corpus to simulate harmful fine-tuning scenarios.
Following the experimental settings used in prior work~\cite{huang2024lisa, yang2025asft}, we set $n=5000$ and $p=0.1$ (4500 downstream task data and 500 unsafe data) for our main results.

\BfPara{Models}
To evaluate the effectiveness of \Sysacro, we use {Qwen2.5-7B-Instruct}~\cite{team2024qwen2}, {Gemma-2-9b-it}~\cite{team2024gemma}, {Llama-3.1-8b-Instruct}~\cite{dubey2024llama} and {Llama-3.2-1b-Instruct}~\cite{touvron2023llama}.
We first conduct a preliminary evaluation of {Qwen2.5-7B-Instruct} to assess the effectiveness of \Sysacro across three datasets.
We then scale the model size as {Llama-3.1-8b-Instruct}, {Gemma-2-9b-it}, and {Llama-3.2-1b-Instruct} to demonstrate that our approach can be applied to larger models, different model families and remains robust as model capacity increases.

\BfPara{Metrics}
Following prior work~\cite{yang2025asft,huang2024lisa}, we evaluate models using fine-tuning accuracy ACC (The top-1 accuracy on the test sets of downstream tasks) and attack successful rate(ASR), which measure downstream task performance and adherence to safety alignment under harmful prompts, respectively.
To ensure a reliable and comprehensive evaluation of safety behavior, we measure safety rates across multiple widely used benchmarks, including StrongREJECT (SR)~\cite{souly2024strongreject}, JailbreakBench (JB)~\cite{chao2024jailbreakbench}, and HarmBench (HB)~\cite{mazeika2024harmbench}.
For all safety evaluations, we employ \texttt{Llama-Guard-3-8B} as the judge model to determine the safety of the model's response and calculate the harmful score ASR.

\BfPara{Baselines}
We compare \Sysacro with 6 baselines: Initial model, vanilla supervised fine-tuning (SFT), SafeLoRA~\cite{hsu2024safe}, Constrained SFT~\cite{qi2024safety}, and AsFT~\cite{yang2025asft}.
We employ LoRA~\cite{hu2022lora} for efficient fine-tuning of all the models, with a rank of 8 across all experiments. The AdamW optimizer is used with a learning rate of \num{3e-5}, training for 3 epochs with a batch size of 2. The coefficient of our PACT regularization term $\lambda_{\mathrm{KL}}$ is set to 3.

\subsection{Main results}

We evaluate \Sysacro across three dimensions: generalization to different fine-tuning datasets, performance across model architectures and sizes, and robustness to varying proportions of harmful data.

\BfPara{Generalization across fine-tuning datasets} 
Table~\ref{tab:different dataset} presents the results for {Qwen-2.5-7B-Instruct} fine-tuned on three diverse tasks, each with 10\% harmful data.
We can see that vanilla SFT consistently achieves high task accuracy but suffers significant safety degradation across all datasets.
For example, the ASRs on the HB dataset are consistently higher than 94\% across all models.
This confirms that even a small proportion of harmful data ($n{=}500$) can substantially compromise model safety.
In such cases, the models lose the direct refusal ability completely.
Existing defense methods exhibit various trade-offs.
SafeLora preserves some safety on GSM8K, but fails completely on SST2 and AGNEWS.
Constrained SFT maintains safety on all the datasets, but they substantially degrade task performance on GSM8K.
AsFT provides a better utility–safety trade-off than Constrained-SFT and Safe LoRA, but still shows elevated jailbreak and harmful behavior rates, particularly on SST2. 
In contrast, \Sysacro consistently achieves the lowest ASRs across all three datasets while maintaining comparable task accuracy.
For example, we attain 80.89\% accuracy on GSM8K while reducing ASR on HB to 29.50\%.
With only marginal accuracy trade-offs, we can approach almost the same safety level of the initial model.
These results demonstrate effective generalization across diverse task domains (mathematical reasoning, sentiment classification, and topic classification) under data poisoning scenarios.

\BfPara{Generalization across model architectures.}
To assess generalization across model architectures and scales, we further evaluate on {Llama-3.1-8B-Instruct}, {Llama-3.2-1B-Instruct}, and {Gemma-2-9B-it}, all fine-tuned on GSM8K with $10\%$ harmful data.
The results are shown in Table~\ref{tab:different_models}.
We can see that existing baselines exhibit inconsistent behavior across models.
SafeLoRA preserves safety on {Llama-3.1-8B} but fails on {Gemma-2-9B}, while AsFT achieves low risk on {Llama-3.2-1B} but catastrophically degrades on {Llama-3.1-8B}.
Such inconsistency indicates that these defenses are sensitive to model architectures and sizes. 
In contrast, \Sysacro consistently achieves the best or near-best safety metrics across all 3 models without sacrificing utility.
Notably, for model {Llama-3.1-8B}, ASR on HB is reduced to 0.00\%, while accuracy remains comparable to SFT. 
On {Gemma-2-9B}, most baselines suffer substantial safety degradation, but \Sysacro constraints the ASR on HB at 14.50\% and achieves the highest task accuracy.
These results confirm that our method generalizes reliably across diverse model families and scales without architecture-specific tuning.

\BfPara{Robustness to harmful data proportions.}
Table~\ref{tab:qwen_generalization} shows the performance under varying poison ratios on the AGNEWS dataset.
We can see that the baseline methods exhibit sharp safety degradation as the poison ratio increases.
For example, SafeLoRA maintains safety at $P=0\%$ but collapses at $P=5\%$ (SR: 71.57\%, HB: 93.50\%). 
AsFT still exhibits substantial risk at higher poison ratios (HB: 70.50\% at $P=10\%$).
While \Sysacro consistently achieves the lowest safety risk across all poison ratios while maintaining task accuracy. 
Notably, even in the benign setting ($P=0\%$), \Sysacro slightly improves upon the initial model's safety (SR: 0.96\% vs. 1.60\%). 
It shows that \Sysacro can mitigate latent safety risks introduced by the fine-tuning process itself, independent of explicit harmful data. 
As the poison ratio increases to $5\%$ and $10\%$, \Sysacro remains stable, unlike the other existing approaches. 

\begin{figure*}[t]
  \centering
  \begin{subfigure}[b]{0.33\linewidth}
    \centering
    \includegraphics[width=0.9\linewidth]{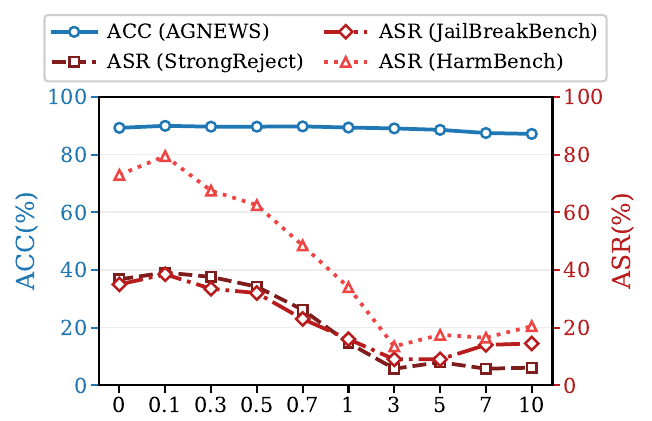}
    \caption{The coefficient ($\lambda_{\mathrm{KL}}$) values}
    \label{fig:ablation:alpha}
  \end{subfigure}%
  \hfill
  \begin{subfigure}[b]{0.33\linewidth}
    \centering
    \includegraphics[width=0.9\linewidth]{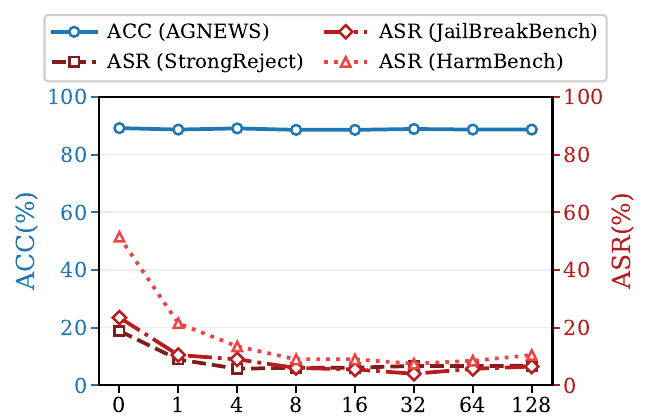}
    \caption{The calibration decay position $N$ values}
    \label{fig:ablation:n1}
  \end{subfigure}%
  \hfill
  \begin{subfigure}[b]{0.33\linewidth}
    \centering
    \includegraphics[width=0.9\linewidth]{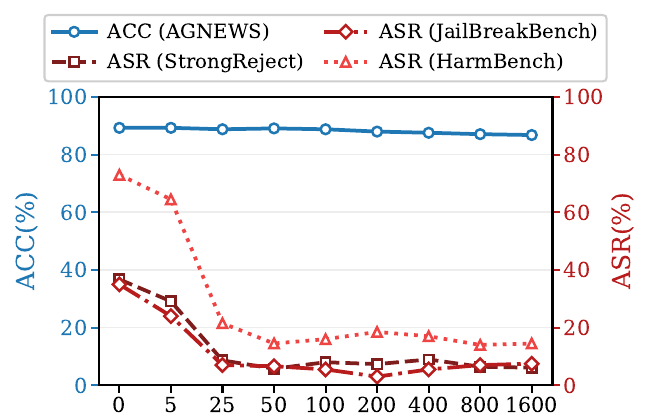}
    \caption{Number of safety tokens $K$.}
    \label{fig:ablation:vtopk}
  \end{subfigure}
  \caption{Hyperparameters sensitivity}
  \label{fig:ablation:four}
\end{figure*}

\subsection{Ablation studies}
We conduct comprehensive ablation studies to validate the contribution of each component in \Sysacro and analyze its sensitivity to key hyperparameters.

\BfPara{Component-wise ablation}
Table~\ref{tab:ablation_singlecol} presents a progressive ablation study on the AGNEWS dataset with 10\% harmful data, where each row incrementally adds one component of \Sysacro to the previous row. 
\begin{table}[h]
\centering
\caption{Ablation study on progressively adding safety-token constraints, each row cumulatively adds the listed constraint(s) from the previous rows.}
\label{tab:ablation_singlecol}
\begin{tabular}{p{2.4cm}cccc}
\toprule
\textbf{Constraints} & \textbf{ACC (\%)}$\uparrow$ & \textbf{SR (\%)}$\downarrow$ & \textbf{JB (\%)}$\downarrow$ & \textbf{HB (\%)}$\downarrow$ \\
\midrule
Full-vocab KL          & 89.30 & 36.74 & 35.00 & 73.00 \\
+Safety tokens  & 88.50 & 25.88 & 28.00 & 69.00 \\
+Safety weights & 89.20 & 18.85 & 23.50 & 51.50 \\
+Response only ref & 88.70 & 6.71  & 5.50  & 10.50 \\
+Position decay      & 89.10 & 5.75  & 9.00  & 13.50 \\
\bottomrule
\end{tabular}
\end{table}
The first row establishes a baseline using standard cross-entropy loss with full-vocabulary KL regularization, where the fine-tuned model is constrained to match the reference model's distribution across all tokens.
The second row (+Safety tokens) restricts the KL regularization to only the identified safety tokens rather than the full vocabulary.
The third row (+Safety weights) further incorporates importance weighting based on the discrepancy score $d(v)$.
The fourth row (+Response only ref) calibrates the full-context reference outputs by mixing them with no-prompt reference distributions.
The last row (+Position decay) adds a position-wise decay coefficient that gradually reduces the weight over the no-prompt outputs.

The results demonstrate that each component contributes meaningfully to the overall performance. 
We observe that restricting KL regularization to safety tokens reduces ASRs across all datasets, demonstrating that focusing regularization on safety-related tokens is substantially more effective than uniform constraints.
Integrating importance weights further reduces ASRs, which proves that they capture the reference model's preference shifts when facing harmful prompts.
The most substantial safety gain comes from incorporating response-only reference model logits.
This confirms that mitigating the influence of harmful prefixes during teacher forcing is critical to maintaining the quality of safety-token supervision.
Finally, adding position-wise decay provides a better trade-off between model utility and safety preservation.

\BfPara{Hyperparameter sensitivity}
We analyze the sensitivity of \Sysacro to three key hyperparameters: the KL regularization coefficient $\lambda_{\mathrm{KL}}$, the calibration decay position $N$, and the number of safety tokens $K$.

Figure~\ref{fig:ablation:four} (a) shows the effect of $\lambda_{\mathrm{KL}}$.
We can see that as it increases, the ASR decreases substantially while task accuracy remains stable. 
We can achieve the optimal safety when $\lambda_{\mathrm{KL}}=3$. 
However, beyond $\lambda_{\mathrm{KL}}=5$, we observe over-regularization, where excessive constraints interfere with downstream task learning.
Nonetheless, \Sysacro can operate effectively within a broad range of $\lambda_{\mathrm{KL}}$, providing flexibility in parameter tuning.

Parameter $N$ controls how much we rely on response-only versus full-context reference distributions across the response.
Figure~\ref{fig:ablation:four} (b) shows that as $N$ increases from 0, ASR decreases while accuracy remains stable.
We can also see that between $N$=8 and $N=128$, \Sysacro shows a stable safety performance. 
This finding aligns with our observation that refusal prefixes (e.g., 'I cannot assist') are generated within the first 5-10 tokens.
On the other hand, applying calibration too deep into the response becomes unnecessary and may introduce noise.

Finally, $K$ determines how many high-discrepancy tokens are constrained during fine-tuning.
Figure~\ref{fig:ablation:four} (c) shows an interesting pattern where models achieve optimal performance at $K = 50$ and maintain relatively stable safety through $K = 1600$.
This provides empirical support for our core insight that safety alignment is remarkably concentrated.
Constraining only 50 tokens is sufficient to prevent alignment drift. 
The additional tokens contribute minimal safety value while potentially introducing unnecessary constraints. 

\begin{figure}[h]
  \centering
  \centering
  \includegraphics[width=0.9\linewidth]{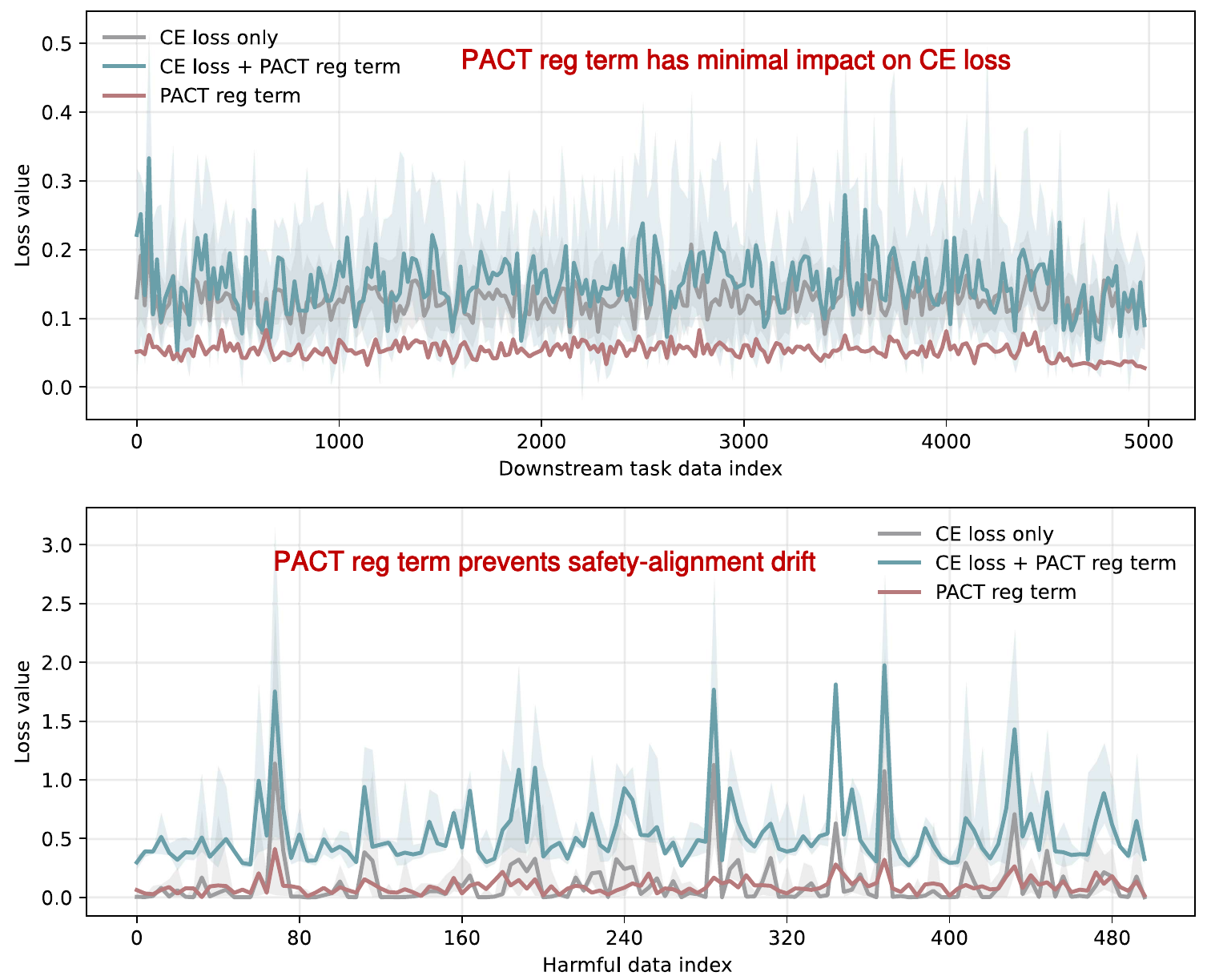}
  \caption{Loss changed during finetuning}
  \label{fig:loss_fig}
\end{figure}

\BfPara{Training dynamics analysis}
To gain deeper insight into how our regularization term interacts with training dynamics, we visualize per-sample loss on both downstream-task and harmful data during training, as shown in Figure~\ref{fig:loss_fig}.
We can see that on downstream task data (top), the CE loss with our regularization term closely tracks the CE loss without it, indicating that our safety-oriented term introduces minimal interference with utility learning.
Moreover, our term itself remains small and stable throughout, confirming that our method imposes negligible constraints on benign samples.

On harmful data (bottom), a markedly different pattern emerges. 
Our KL regularization term exhibits substantially larger values than those on downstream task data, demonstrating that it selectively activates on harmful samples and effectively distinguishes between benign and harmful data without explicit labels.
Importantly, the CE loss under our regularization term is consistently higher than the unregularized CE loss throughout training.
This shows that our term actively resists the model's attempt to fit harmful responses by anchoring it to the reference model's safe behavior, thereby slowing the erosion of safety alignment.
Meanwhile, our regularization term aims to preserve safety behaviors throughout the fine-tuning.

\begin{figure}[h]
  \centering
  \centering
  \includegraphics[width=0.9\linewidth]{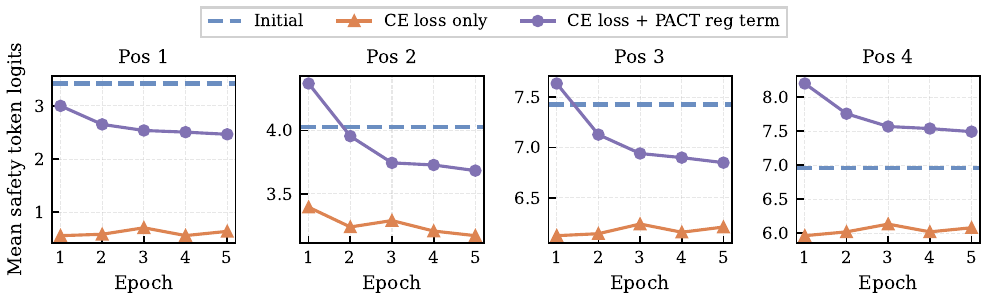}
  \caption{Mean safety token logits on first 4 answer position}
  \label{fig:mean_logit}
\end{figure}

\BfPara{Token-level analysis}
To further understand the mechanism by which \Sysacro preserves safety, we track the mean logits of the selected safety tokens at the first 4 answer positions across 5 training epochs, as shown in Figure~\ref{fig:mean_logit}. 
These positions are particularly critical, as they correspond to where typical refusal prefixes are generated. 
Under standard CE loss-only training, the safety token logits collapse rapidly and remain far below the initial model's values across all 4 positions, indicating that SFT quickly suppresses the model's refusal behavior. 
In contrast, with our regularization term, the safety token logits are substantially better preserved throughout training, staying much closer to the initial model's distribution. 
Although a gradual decline is still observed, which reflects the inherent tension between task adaptation and safety preservation, the gap relative to the initial model is significantly narrower than under CE loss-only training. 
This confirms that \Sysacro effectively anchors the safety-critical output distribution at the most relevant safety tokens and answer positions, preventing the rapid logit drift that leads to safety degradation.

\section{Related works}

\BfPara{Alignment preserving fine-tuning}
To mitigate alignment drift introduced during fine-tuning, existing approaches can be broadly categorized into two groups: parameter-level and data-level safety preservation. For parameter-level preservation, 
SafeLoRA~\cite{hsu2024safe} constrains LoRA updates by projecting them onto a safety-aligned subspace induced by aligned–unaligned weight differences, while ASFT~\cite{yang2025asft} preserves alignment via an aligned–misaligned weight-difference regularizer that keeps updates within a safety basin. 
SALoRA~\cite{li2025salora} enables safety-aware adaptation by pruning alignment-irrelevant parameters and updating only those most relevant to safety.
Antidote~\cite{huang2024antidote} effectively prunes harmful parameters after fine-tuning.
SOMF~\cite{sheshadri2024latent} integrates utility-relevant knowledge from benign tasks while preserving safety-critical parameters.
NLSR~\cite{yi2025nlsr} designs a training-free preserving that restores safety after fine-tuning by patching safety-critical parameters identified via pre/post similarity shifts. 
For data-level preservation, 
SAFT~\cite{choi2024safety} filters harmful data using subspace-decomposition scores. SafeInstr~\cite{bianchi2023safety} incorporates safety-focused examples to enhance safety behavior during instruction tuning.
Lisa~\cite{huang2024lisa} constrains optimization drift via dual-state training with alignment data and proximity regularization. 
BEA~\cite{wang2024mitigating} embeds covert triggers to suppress harmful outputs at inference time.


\BfPara{Token-level effects on alignment}
Recent studies~\cite{lin2023unlocking,fei2025nudging} suggest that alignment changes can be concentrated in a small set of localized signals, such as specific tokens or short patterns that govern refusal style, formatting, or safety-related phrasing, rather than requiring broad redistribution over the entire vocabulary. Meanwhile, ~\cite{qi2024safety} introduces shallow safety alignment, which finds the tendency for safety to concentrate on the first few output tokens. ~\cite{jain2024refusal} proposes category-conditioned refusal tokens and steers per-category refusal rates at inference by modulating refusal-token likelihood, avoiding retraining.

\section{Conclusion}
In this work, we study safety-alignment drift as an intrinsic risk of downstream fine-tuning, showing that even benign training data can weaken refusal behaviors, and that small amounts of harmful data can further amplify this failure mode. To mitigate this, we introduce \Sysacro, a token-level safety-preserving fine-tuning framework that stabilizes the model’s output confidence on a compact set of safety-critical tokens. By regularizing the fine-tuned model to match a safety-aligned reference model’s per-step confidence only on safety-related tokens, while leaving the remaining token space largely unconstrained, \Sysacro prevents refusal degradation without imposing model-wide restrictions that often harm downstream utility. 
Overall, our results demonstrate that fine-grained safety alignment control is achievable at the token level. 
\Sysacro improves robustness to safety alignment drift while maintaining effective task adaptation during fine-tuning.

\section{Acknowledgments}
{This work has been supported by an ONR grant N00014-23-1-2137 and an NSF award CNS-2442976.}

\bibliographystyle{ACM-Reference-Format}
\bibliography{ref}

\appendix
\section{Additional Results and Clarifications}
\label{app:additional-results}

This appendix provides additional results and clarifications that complement the main paper.

\subsection{More Comprehensive Baseline Methods}
\label{app:more-baselines}

We added three stronger baselines: SAFEGRAD, SafeDelta, and Lisa, under the setting of AGNEWS, $P=10\%$, Llama-3.1-8B-Instruct. Compared with these new baselines, PACT not only achieves substantially higher ACC but also attains clearly lower ASR on HB. The new results Table~\ref{tab:app_more_baselines}, evaluated by Llama-Guard-3-8B, show that PACT still achieves the best overall accuracy-safety trade-off.

\begin{table}[h]
\centering
\caption{Comparison with additional baseline methods under AGNEWS, $P=10\%$, and Llama-3.1-8B-Instruct. The results are evaluated by Llama-Guard-3-8B.}
\label{tab:app_more_baselines}
\begin{tabular}{lrrrr}
\toprule
Methods & ACC & SR & JB & HB \\
\midrule
Initial & 75.90 & 0.32 & 3.00 & 9.50 \\
SAFEGRAD & 87.20 & 1.32 & 2.00 & 9.00 \\
SafeDelta & 77.20 & 1.64 & 1.50 & 9.00 \\
Lisa & 79.40 & 0.96 & 3.50 & 13.50 \\
\textbf{PACT (Ours)} & \textbf{87.80} & 2.56 & \textbf{1.50} & \textbf{1.00} \\
\bottomrule
\end{tabular}
\end{table}

\subsection{Higher Proportions of Harmful Data}
\label{app:higher-harmful-ratios}

As shown as Table~\ref{tab:app_higher_harmful_ratios}, we extend the experiments to higher harmful data ratios, $P=20\%$ and $P=30\%$. Specifically, under the Qwen2.5-7B-Instruct model with the AGNEWS dataset fine-tuning setting, we augment the original 500 AdvBench harmful samples with an additional 500 and 1000 harmful samples from BeaverTails, respectively.

\begin{table*}[t]
\centering
\caption{Results under higher harmful data ratios, $P=20\%$ and $P=30\%$, using Qwen2.5-7B-Instruct on AGNEWS.}
\label{tab:app_higher_harmful_ratios}
\begin{tabular}{lrrrrrrrr}
\toprule
Methods 
& ACC ($P=20\%$) & SR ($P=20\%$) & JB ($P=20\%$) & HB ($P=20\%$)
& ACC ($P=30\%$) & SR ($P=30\%$) & JB ($P=30\%$) & HB ($P=30\%$) \\
\midrule
Initial & 83.20 & 1.60 & 3.00 & 7.00 & 83.20 & 1.60 & 3.00 & 7.00 \\
Ours & 89.30 & 2.88 & 6.50 & 10.50 & 89.10 & 3.51 & 5.50 & 11.00 \\
\bottomrule
\end{tabular}
\end{table*}

This result demonstrates that PACT is able to consistently preserve the model's safety under more challenging attack scenarios, while still maintaining the accuracy of the fine-tuned model.

\subsection{Alternative Evaluators}
\label{app:alternative-evaluators}

As shown as Table~\ref{tab:app_beaverdam_eval}, we additionally evaluated PACT and the newly added baselines using Beaver-Dam-7B~\cite{ji2023beavertails} under the AGNEWS, $P=10\%$, Llama-3.1-8B-Instruct experiment setting.

\begin{table}[h]
\centering
\caption{Evaluation with Beaver-Dam-7B under AGNEWS, $P=10\%$, and Llama-3.1-8B-Instruct.}
\label{tab:app_beaverdam_eval}
\begin{tabular}{lrrrr}
\toprule
Methods & ACC & SR & JB & HB \\
\midrule
Initial & 75.90 & 2.56 & 11.00 & 5.00 \\
SAFEGRAD & 87.20 & 1.92 & 11.50 & 5.50 \\
SafeDelta & 77.20 & 1.92 & 13.50 & 5.50 \\
Lisa & 79.40 & 1.92 & 12.50 & 5.50 \\
\textbf{PACT (Ours)} & \textbf{87.80} & 2.88 & \textbf{7.00} & \textbf{1.00} \\
\bottomrule
\end{tabular}
\end{table}

The conclusions are consistent with those using Llama-Guard-3-8B: PACT still achieves the best average safety performance among all baselines. This demonstrates that our results are robust to the choice of evaluator.

\subsection{Overall Generalization of Safety Tokens}
\label{app:safety-token-generalization}

Llama-3.1, Llama-3.2, and Gemma-2 do not reuse the safety tokens identified from Qwen2.5. For each model, we re-identify safety tokens using the discrepancy-based procedure in Sec.~3.1. Examples of the top-10 safety tokens in each of these three models are shown below.

For Gemma-2-9B-it, the top-10 safety tokens are:
\begin{quote}
``I'', ``cannot'', ``and'', ``*'', ``your'', ``is'', ``purpose'', ``helpful'', ``fulfill'', and ``request''.
\end{quote}

For Llama-3.1-8B-Instruct, the top-10 safety tokens are:
\begin{quote}
``can'', ``I'', ``'t'', ``that'', ``request'', ``with'', ``fulfill'', ``is'', ``$<$|eot\_id|$>$'', and ``assist''.
\end{quote}

For Llama-3.2-1B-Instruct, the top-10 safety tokens are:
\begin{quote}
``I'', ``can'', ``'t'', ``request'', ``that'', ``fulfill'', ``assist'', ``with'', ``or'', and ``provide''.
\end{quote}

\subsection{Computational and Implementation Cost}
\label{app:computational-cost}

As shown as Table~\ref{tab:app_computational_cost}, we report the computational overhead in our main experimental setting using Llama-3.1-8B-Instruct and AGNEWS with $P=10\%$. This engineering trade-off is discussed more clearly in the revision.

\begin{table}[t]
\centering
\caption{Computational overhead under Llama-3.1-8B-Instruct and AGNEWS with $P=10\%$. C-SFT denotes Constrained-SFT and V-SFT denotes Vanilla SFT.}
\label{tab:app_computational_cost}
\small
\setlength{\tabcolsep}{3pt}
\begin{tabular}{lcccc}
\toprule
Metric & C-SFT & V-SFT & PACT & Ratio \\
\midrule
Train time (s) & 1589.9 & 1071.34 & 1710.68 & 1.60$\times$ \\
Step time (s) & 0.1893 & 0.1417 & 0.2267 & 1.60$\times$ \\
Memory (GB) & 21.86 & 18.68 & 33.95 & 1.82$\times$ \\
\bottomrule
\end{tabular}
\end{table}

\subsection{Causality and Adversarial Robustness}
\label{app:causality-adversarial-robustness}

We explicitly account for adversarial contamination during teacher forcing via the calibration mechanism in Sec.~3.2. We also added a stronger stress test. Under AGNEWS, $P=10\%$, and Llama-3.1-8B-Instruct, we additionally injected highly adversarial prompts from Persuasive-Jailbreaker-Data into the fine-tuning data. The results are shown in Table~\ref{tab:app_pjd_training}.

\begin{table}[h]
\centering
\caption{Injecting highly adversarial prompts from Persuasive-Jailbreaker-Data into the fine-tuning data under AGNEWS, $P=10\%$, and Llama-3.1-8B-Instruct.}
\label{tab:app_pjd_training}
\begin{tabular}{lrrrr}
\toprule
Methods & ACC & SR & JB & HB \\
\midrule
SFT & 89.50 & 96.81 & 54.50 & 95.50 \\
Ours & 87.60 & 2.88 & 2.50 & 1.50 \\
\bottomrule
\end{tabular}
\end{table}

This shows that even when the attacker targets the training data with substantially stronger adversarial prompts, PACT still preserves a strong utility-safety trade-off.

\vfill\eject
\subsection{Comparison with a Simple Data-Centric Baseline}
\label{app:simple-data-centric-baseline}

We added 500 safe examples from PKU-SafeRLHF-QA~\cite{ji2025pkusaferlhf} to the original AGNEWS $P=10\%$ setting, forming a simple data-centric baseline, and also included Lisa. The results are shown in Table~\ref{tab:app_safe_data_baseline}.

\begin{table}[h]
\centering
\caption{Comparison with a simple data-centric baseline under AGNEWS, $P=10\%$, and Llama-3.1-8B-Instruct.}
\label{tab:app_safe_data_baseline}
\begin{tabular}{lrrrr}
\toprule
Methods & ACC & SR & JB & HB \\
\midrule
Add extra 10\% safe data & 90.80 & 15.65 & 46.00 & 84.50 \\
Lisa & 79.40 & 0.96 & 3.50 & 13.50 \\
Ours & 87.80 & 2.56 & 1.50 & 1.00 \\
\bottomrule
\end{tabular}
\end{table}

\subsection{Adversarial Prompts with Refusal-Like Prefixes}
\label{app:refusal-like-prefixes}

We created a stronger setting by prepending a fixed refusal prefix, ``I'm sorry, I can't help you.'', to PJD, yielding Refusal-PJD, where later content can still become harmful. Mixing these prompts into AGNEWS $P=10\%$ gives the results in Table~\ref{tab:app_refusal_pjd}.

\begin{table}[h]
\centering
\caption{Results under adversarial prompts with refusal-like prefixes.}
\label{tab:app_refusal_pjd}
\begin{tabular}{lrrrr}
\toprule
Methods & ACC & SR & JB & HB \\
\midrule
SFT-PJD & 89.50 & 96.81 & 54.50 & 95.50 \\
PACT-PJD & 87.60 & 2.88 & 2.50 & 1.50 \\
SFT-Refusal-PJD & 89.50 & 96.49 & 57.50 & 95.00 \\
PACT-Refusal-PJD & 87.60 & 3.83 & 4.00 & 5.50 \\
\bottomrule
\end{tabular}
\end{table}

PACT does not simply relax once it sees refusal-like tokens; it remains substantially more robust even when harmful content appears later under misleading refusal prefixes.

\end{document}